\pdfoutput=1

\documentclass[11pt]{article}

\usepackage{acl}

\usepackage{times}
\usepackage{latexsym}

\usepackage[T1]{fontenc}

\usepackage[utf8]{inputenc}

\usepackage{microtype}

\usepackage{inconsolata}

\usepackage{graphicx}
\usepackage[ruled,linesnumbered]{algorithm2e}
\SetKwComment{Comment}{$\triangleright$\ }{}
\usepackage{amsmath}
\usepackage{bbding}
\usepackage{booktabs}
\usepackage{multirow}
\usepackage{amsthm}

\title{Multi-Candidate Speculative Decoding}

\author{Sen Yang,\ Shujian Huang{\thanks{\ \ Corresponding author.}},\ Xinyu Dai,\ Jiajun Chen \\
  National Key Laboratory for Novel Software Technology, Nanjing University \\
  \texttt{yangsen@smail.nju.edu.cn}\\\texttt{\{huangsj,daixinyu,chenjj\}@nju.edu.cn}}

\begin{document}
\maketitle
\begin{abstract}
Large language models have shown impressive capabilities across a variety of NLP tasks, yet their generating text autoregressively is time-consuming. One way to speed them up is speculative decoding, which generates candidate segments (a sequence of tokens) from a fast draft model that is then verified in parallel by the target model. However, the acceptance rate of candidate tokens receives limitations from several factors, such as the model, the dataset, and the decoding setup. This paper proposes sampling multiple candidates from a draft model and then organising them in batches for verification. We design algorithms for efficient multi-candidate verification while maintaining the distribution of the target model.
Our approach shows significant improvements in acceptance rates on multiple datasets and models, consistently outperforming standard speculative decoding.\footnote{We release our code, datasets, and model checkpoints at \url{https://github.com/NJUNLP/MCSD}.}
\end{abstract}


\section{Introduction}


Recently developed large language models (LLMs), such as GPT series~\cite{brown2020language, achiam2023gpt} and LLaMA~\cite{touvron2023llama_a, touvron2023llama_b}, have demonstrated remarkable capabilities in language understanding and generation, as well as generalizability across a wide variety of NLP tasks and open domains. This promotes the need to deploy LLM services. However, the extensive volume of parameters and computational overhead make LLMs run with significantly higher latency than smaller models.
On the other hand, popular Transformer-based LLMs typically generate text in an autoregressive paradigm, which necessitates multiple iterations of the model for decoding a single piece of text, making things even worse.

To reduce the inference cost, a series of methods have have been developed~\cite{so2021primer, shazeer2019fast, jaszczur2021sparse, kwon2023efficient}. Among them, \emph{speculative decoding} (SD)~\cite{leviathan2023fast, chen2023accelerating} has been proved to be very effective in improving the end-to-end latency of large autoregressive models without compromising the quality of generation. SD employs an additional draft model, typically much smaller than the target model to be accelerated, to generate a few candidate tokens at low computational cost. These candidates are then verified in parallel on the target model. 

The main purpose of SD is to minimize the invocations of the target model by accepting as many tokens as possible during the verification stage. Therefore, the acceleration performance critically depends on the acceptance rate of candidate tokens by the target model, i.e., the agreement between the draft and target model’s distributions under a given context.
In general, models within the same suite generally exhibit strong agreement in their output distributions. However, our experiments have revealed that the distributional discrepancies between the target and draft models become more pronounced when tackling complex tasks that involve longer prompts. On the other hand, it has become popular in the community to fine-tune LLMs using additional data to enhance their performance in specific aspects~\cite{vicuna2023, alpaca, iyer2022opt, chung2022scaling}. It is important to note that fine-tuning can also introduce significant distributional discrepancies between the target and draft models, even if they were initially well-aligned.

The aim of this work is to improve the acceptance rate of the target model during the verification stage. Our motivation is to sample multiple candidate tokens at each position in the draft generation. These candidates are organized in a batch for parallel verification on the target model. 
Although this approach is straightforward and intuitive, it encounters the challenge that SD cannot directly utilize multiple candidates to improve the acceptance rate while preserving the output distribution of the target model. To address it, we propose the algorithm for multi-candidate verification. Moreover, the multiple candidates sampled from the draft model have a probability of collision. Thus, we also introduce a more efficient version for candidates sampled without replacement.




We evaluate our method using the LLaMA suite, including its fine-tuned version, Vicuna, with both argmax and standard sampling. Our method yields significant improvements in acceptance rates on the Alpaca and WMT datasets, consistently outperforming SD in walltime speed. We further validate our method's generalizability across models by extending it to the LLaMA2 and OPT suites. Notably, acceptance rates can often be improved by fine-tuning the draft model, and we show that our method can be superimposed on it.

\section{Background: Speculative Decoding}
\label{sec:sd}

The workflow of speculative decoding is shown in Fig.~\ref{fig:ARCH}(a). Given contexts $c$, speculative decoding starts by invoking the draft model $M_q$ to sample a draft sequence of tokens with a length of $\gamma$, denoted as $\tilde{x}_{1}, \cdots, \tilde{x}_{\gamma}$, where $\tilde{x}_{i}\sim q(x|\tilde{x}_1,\cdots,\tilde{x}_{i-1},c)$. The draft tokens, along with the contexts, are then passed to the target model $M_p$ to obtain their output distribution $p(x|\tilde{x}_1,\cdots,\tilde{x}_i,c)$ in parallel. Finally, the draft tokens is verified sequentially from $\tilde{x}_1$ to $\tilde{x}_{\gamma}$. To verify token $\tilde{x}_i$, a speculative sampling algorithm is employed to determine whether to accept $\tilde{x}_i$ or not, based on $q(x|\tilde{x},\cdots,\tilde{x}_{i-1},c)$ and $p(x|\tilde{x},\cdots,\tilde{x}_{i-1},c)$. Once a token is rejected, the next verification terminates and the algorithm returns a new token as the endpoint. If all tokens are accepted, there is an extra token sampled from $p(x|\tilde{x}_1,\cdots,\tilde{x}_\gamma,c)$ as the endpoint. Thus, the process generates a minimum of $1$ and a maximum of $\gamma+1$ accepted tokens.

\paragraph{Speculative Sampling.} The significance of speculative sampling is that we cannot accept the guesses given by the draft model without restriction, otherwise we cannot maintain the same output distribution as the target model. A simple and straightforward idea is to first sample a token $x$ from $p(x)$\footnote{We'll use $p(x)$ and $p(\tilde{x})$ to denote $p(x|\tilde{x}_1,\cdots,\tilde{x}_{i-1}, c)$ and $p(\tilde{x}_i|\tilde{x}_1,\cdots,\tilde{x}_{i-1}, c)$ respectively whenever the prefix is clear from the context, and similarly for $q(x)$ and $q(\tilde{x})$.}, accept $\tilde{x}$ if $x=\tilde{x}$, otherwise reject it and return $x$. However, this approach --- what we call \emph{naive sampling} --- has a very inefficient acceptance rate of $\sum_{\tilde{x}} p(\tilde{x})q(\tilde{x})$. As a comparison, speculative sampling uses a novel algorithm that accepts $\tilde{x}$ with probability $\min(1, \frac{p(\tilde{x})}{q(\tilde{x})})$, leading to an overall acceptance rate of $\sum_{\tilde{x}} \min(p(\tilde{x}), q(\tilde{x}))$. If $\tilde{x}$ is rejected, then return a new token sampling from $p^\prime (x)=\frac{\max(0, p(x)-q(x))}{\sum_x \max(0, p(x)-q(x))}$. It can be proven that speculative sampling can maintain the same output distribution as the target model\cite{leviathan2023fast, chen2023accelerating} while possessing an upper bound on the acceptance rate of naive sampling (Appendix~\ref{sec:proof_NSUB}).


\section{Multi-Candidate Speculative Decoding}

\begin{figure}[ht!]
    \centering
    \includegraphics[scale=1.05]{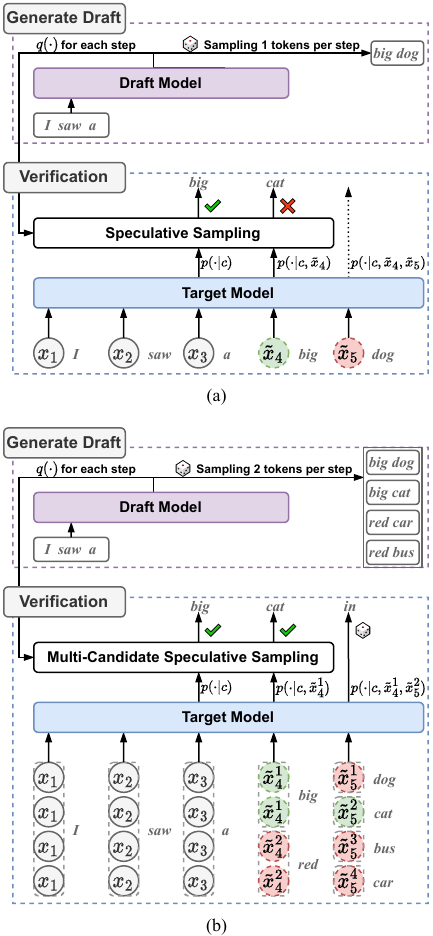}
    \vspace{-3ex}
    \caption{The procedure for standard SD~(a) and MCSD~(b).}
    \vspace{-2ex}
    \label{fig:ARCH}
\end{figure}

Behind the success of SD lies the effective utilization of parallel computing on computing devices: the latency of parallel scoring of short continuations is comparable to that of sampling a single token from the larger target model.
Ideally, the length of draft tokens generated by the draft model (i.e., $\gamma$) can be increased all the way up to the upper limit of computing devices.
However, there comes the diminishing marginal utility with an increase in $\gamma$ rapidly, as the acceptance of a draft token for a given position depends not only on itself but also on the acceptance of all preceding tokens.

In short, there is a portion of potential computing resources that have not been fully utilized. Our work involves utilizing this portion of resources to perform parallel verification on another dimension (i.e., the batch dimension) to significantly improve the acceptance rate of draft tokens. This requires the draft model to sample more than one token at each step, eventually generating a batch of candidates. This batch of candidates is fed together into the target model to obtain the output distribution at each position, as shown in Fig.~\ref{fig:ARCH}(b).

The verification process for multiple candidates is roughly the same as SD: starting from the first step of generation, input the output distributions $q, p$ and candidate tokens $\tilde{x}^1,\cdots,\tilde{x}^k$ for the current step into the speculative sampling algorithm. If the algorithm accepts one of the $k$ tokens, the candidate corresponding to that token is retained in the batch for the next step of verification. If the tokens are all rejected, the algorithm returns a new token as the endpoint, and the next verification procedure is aborted. Finally, if a candidate in the batch survives to the end, the endpoint token is sampled from the output distribution corresponding to this candidate.
Taking a look at the process, it is similar to walking from the root of a tree to a leaf node, where each step chooses a path from one of the k branches or aborts early.

\subsection{Multi-Candidate Speculative Sampling}
\label{sec:MCSSwR}

Now the problem is that the original speculative sampling algorithm described in Section~\ref{sec:sd} cannot be directly used for verification of multiple candidates. This motivates us to design the speculative sampling algorithms for multi-candidate verification, as shown in Algorithm \ref{algorithm:MCSS}. In Appendix~\ref{sec:proof_MCSS}, we prove that the tokens output from Algorithm~\ref{algorithm:MCSS} follow the distribution $p(x)$, which ensures that the output obtained by our algorithm has the same distribution as the target model.

\begin{algorithm}[htb]
\caption{Multi-Candidate Speculative Sampling}
\label{algorithm:MCSS}

\KwIn{Distributions $p,q$; The candidate tokens $\tilde{x}^1,\cdots,\tilde{x}^k\sim q$. }
\KwOut{If accept, returns \emph{True} and the accepted token, otherwise returns \emph{False} and a new token as the endpoint.}
\For{$i\leftarrow 1$ \KwTo $k$}{
    $r\sim U(0,1)$\BlankLine
    \uIf(\textcolor{blue}{\Comment*[f]{Accept $\tilde{x}^i$}}){$r\leq \frac{p(\tilde{x}^i)}{q(\tilde{x}^i)}$}{
        \Return{$(\mathit{True}, \tilde{x}^i)$}
    }\uElse{
        \textcolor{blue}{\Comment{Normalize the residual $p(x)$}}
        $p(x):=\frac{\max(0, p(x)-q(x))}{\sum_x \max(0, p(x)-q(x))}$
    }
}
$x_{\mathrm{end}}\sim p(x)$\BlankLine
\Return{$(\mathit{False}, x_{\mathrm{end}})$}
\end{algorithm}

\subsection{Multi-Candidate Speculative Sampling without Replacement}
\label{sec:MCSSwoR}

\begin{algorithm}[htb]
\caption{Multi-Candidate Speculative Sampling without Replacement}
\label{algorithm:MCSSwoR}

\KwIn{Distributions $p,q$; The candidate tokens $\tilde{x}^1,\cdots,\tilde{x}^k$, where $\tilde{x}^i\sim \bar{q}_{i-1}(x)$. }
\KwOut{If accept, returns \emph{True} and the accepted token, otherwise returns \emph{False} and a new token as the endpoint.}
\For{$i\leftarrow 1$ \KwTo $k$}{
    $r\sim U(0,1)$\BlankLine
    \uIf(\textcolor{blue}{\Comment*[f]{Accept $\tilde{x}^i$}}){$r\leq \frac{p(\tilde{x}^i)}{q(\tilde{x}^i)}$}{
        \Return{$(\mathit{True}, \tilde{x}^i)$}
    }\uElse{
        \textcolor{blue}{\Comment{Normalize the residual $p(x)$ and $q(x)$}}
        $p(x):=\frac{\max(0, p(x)-q(x))}{\sum_x \max(0, p(x)-q(x))}$\BlankLine
        $q(\tilde{x}^i)\leftarrow 0$\BlankLine
        $q(x):=\frac{q(x)}{\sum_x q(x)}$
    }
}
$x_{\mathrm{end}}\sim p(x)$\BlankLine
\Return{$(\mathit{False}, x_{\mathrm{end}})$}
\end{algorithm}

Notice that the $k$ tokens sampled by the draft model at each step are independently and identically distributed. It is likely that $\tilde{x}^1,\cdots,\tilde{x}^k$ are not unique to each other as $k$ increases. Even if a token is repeatedly sampled, there is no way to increase the probability of accepting it, because once it is rejected the first time, it has a probability of $0$ in the residual distribution $p$, and thus will never be accepted again.

An intuitive way to prevent token collisions is to do sampling without replacement in draft model generation. Without loss of generality, assume that $\tilde{x}^1, \cdots ,\tilde{x}^k$ are sampled from $q$ sequentially, that is $\tilde{x}^i \sim \bar{q}_{i-1}(x)$, where
\begin{equation}
\label{equ:q_bar}
\begin{split}
    \bar{q}_0(x)&=q(x),\\
    \bar{q}_i(x)&=\begin{cases}
    0,& x\in \{\tilde{x}^1,\cdots,\tilde{x}^{i}\}\\ 
    \frac{q(x)}{\sum_{x\neq \tilde{x}^1,\cdots,\tilde{x}^{i}}{q(x)}},& \text{otherwise}
    \end{cases}.
\end{split}
\end{equation}

We also propose the speculative sampling algorithm as shown in Algorithm~\ref{algorithm:MCSSwoR} and the proof that it maintains the distribution $p(x)$ in Appendix~\ref{sec:proof_MCSSwoR}.

\subsection{Tree Attention}
\label{sec:tree_attn}

\begin{figure}[htb]
    \centering
    \includegraphics[scale=0.88]{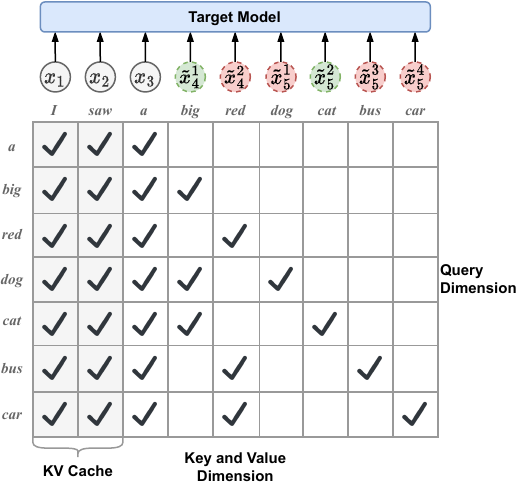}
    \vspace{-3ex}
    \caption{Processing multiple candidates in a single sequence concurrently based on Tree Attention, which contains an attention mask that exclusively permits each token to access its ancestors.}
    \vspace{-2ex}
    \label{fig:tree_attn}
\end{figure}

The generative Transformer~\cite{vaswani2017attention}, which serves as the backbone of LLMs, employs a left-to-right manner in text generation based on causal language modeling. To generate each new token, the attention mechanism requires accessing the keys and values of all preceding tokens. Due to the forward dependency, once a token is generated, the keys and values at that position remain unchanged in the following iterations, so it is very common to cache the keys and values of generated tokens for reuse. In multiple candidate generation and verification, these cached keys and values should be copied within the batch, for both the target and draft models. Although the copying incurs a slight overhead, the situation worsens as these keys and values need to be transferred to the computational unit at each layer of the model, even if they are numerically duplicated.

Here we employ Tree Attention~\cite{miao2023specinfer, medusa} to mitigate the communication overhead resulting from replicated caches, which enables multiple candidates to share the caches of generated tokens. This time, all candidates are squeezed onto a single sequence in the order in which they were generated\footnote{The positional encoding of each token keeps the respective position in the previous sequence unchanged.}. Then, a well-designed attention mask is applied to the sequence to prevent information contamination among candidates and preserve causal relationships between tokens, as shown in Fig.~\ref{fig:tree_attn}.
With multiple candidate sequences arranged in a single sequence, the length of the sequence is slightly increased compared to the original. Accordingly, the total computational overhead is increased, but it is negligible compared to the communication overhead saved, since the contextual prefixes are generally much longer than the length of the candidates generated by the draft model. In our architecture, multiple candidates can be maximally squeezed into a single sequence without adding too much length, thanks to the $k$-ary tree formed by the candidate tokens, which allows a previously generated token to be shared by its descendants in the sequence.


\section{Experiments}

\begin{figure*}[htb]
    \centering
    \includegraphics[scale=0.39]{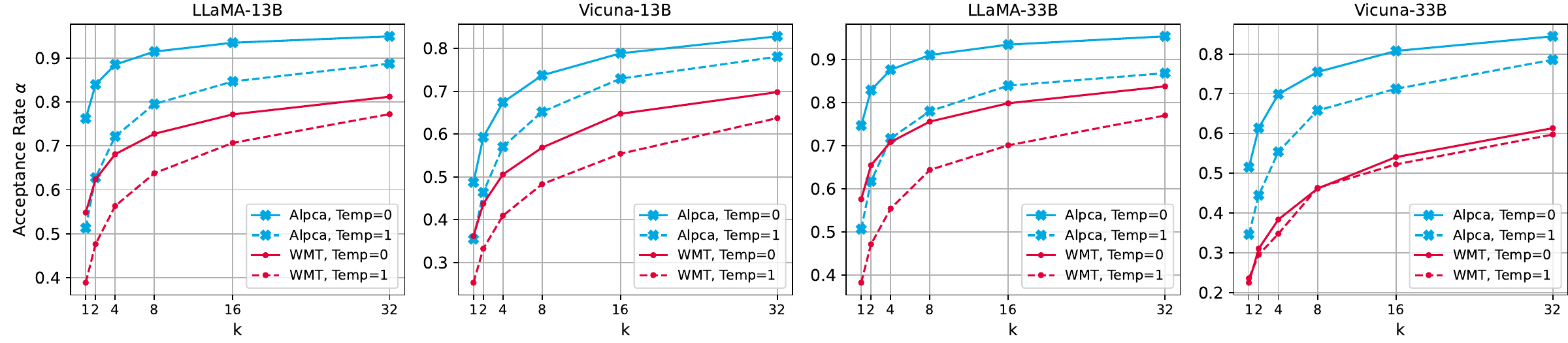}
    \vspace{-1ex}
    \caption{Acceptance rate ($\alpha$) curves given different $k$ using LLaMA-68M as draft model.}
    \vspace{-2ex}
    \label{fig:alphas}
\end{figure*}

\subsection{Experiment Settings}

\paragraph{Models.}

Our evaluation is based on the publicly available LLM suite LLaMA~\cite{touvron2023llama_a}, as well as its fine-tuned version Vicuna~\cite{vicuna2023}, which has been fine-tuned with instruction data to better perform dialogue as well as execute instructions.
We select the 13B and 33B size versions as our target models. 
Since there are no small models suitable for rapid draft generation included in the LLaMA suite, we employ LLaMA-68M and LLaMA-160M\cite{miao2023specinfer} as draft models, which are trained from scratch on the Wikipedia dataset and part of the C4 dataset\cite{raffel2020exploring}.

\paragraph{Datasets.}

We evaluate our approach on the conversational dataset Alpaca~\cite{alpaca, peng2023instruction} and the translation dataset WMT EnDe~\cite{bojar2014findings}, which were used in previous works~\cite{leviathan2023fast, zhou2023distillspec} to evaluate decoding acceleration for LLM, and we observe that the performance of SD varies dramatically between the two datasets. For the WMT dataset, the source text is embedded in an instruction template to prompt the model to perform the translation task.

\paragraph{Efficiency Measures.}

We use the \textbf{acceptance rate $\alpha$} to evaluate the probability that a candidate token is accepted at each step, which basically indicates the distributional consistency between the draft model and the target model.

Since the draft model generates candidate segments of $\gamma$ tokens at a time for verification, the \textbf{block efficiency $\tau$} is commonly used as the expected number of generated tokens per block~\cite{leviathan2023fast}.
Note that in the worst case, $\tau=1$, since the algorithm at least returns a token as the endpoint; if all candidate tokens are accepted, the target model appends an extra token at the end, see Fig.~\ref{fig:ARCH}(b), and so $\tau= \gamma + 1$.

In a real-world deployment, calling the draft model and executing our algorithm incurs additional overheads, so we report average wall clock speedups besides block efficiency.

\paragraph{Platform.}

The experiments were conducted on a single node, of which is equipped with four NVIDIA RTX A6000 48GB GPUs. All systems serves in half precision. The 13B versions are deployed on one GPU and the 33B versions are deployed across two GPUs. The draft models are deployed along with the target models and does not occupy additional GPUs.

\paragraph{Inference Settings.}

Since LLaMA base models are unlikely to stop generating naturally, we limit the generation length of all models to a maximum of 128 new tokens at inference. Our experiments involve two popular sampling setups, argmax sampling (temp=0) and standard sampling (temp=1). For other sampling methods, they can all easily be cast into standard sampling from an adjusted probability distribution.

\subsection{Acceptance Rate Improvement}

\begin{table}[htb]
\footnotesize
\centering
\begin{tabular}{l|l|c|rr}
\toprule
Draft model & Target model & Temp           & Alpaca  & WMT \\ \midrule
\multirow{8}*{\scriptsize LLaMA-68M} & LLaMA-13B  & \multirow{4}*{0} & 0.76    &  0.55 \\
& LLaMA-33B  &                  & 0.75    &  0.58 \\
& Vicuna-13B &                  & 0.49    &  0.36 \\
& Vicuna-33B &                  & 0.52    &  0.22 \\
\cmidrule{2-5}
& LLaMA-13B  & \multirow{4}*{1} & 0.51    &  0.39 \\
& LLaMA-33B  &                  & 0.51    &  0.38 \\
& Vicuna-13B &                  & 0.36    &  0.25 \\
& Vicuna-33B &                  & 0.35    &  0.24 \\
\midrule \midrule
\multirow{8}*{\scriptsize LLaMA-160M} & LLaMA-13B  & \multirow{4}*{0} & 0.80    &  0.59 \\
& LLaMA-33B  &                  & 0.78    &  0.61 \\
& Vicuna-13B &                  & 0.54    &  0.39 \\
& Vicuna-33B &                  & 0.54    &  0.25 \\
\cmidrule{2-5}
& LLaMA-13B  & \multirow{4}*{1} & 0.57    &  0.43 \\
& LLaMA-33B  &                  & 0.57    &  0.42 \\
& Vicuna-13B &                  & 0.42    &  0.29 \\
& Vicuna-33B &                  & 0.42    &  0.25 \\
\bottomrule
\end{tabular}
\caption{Acceptance rate ($\alpha$) on Alpaca and WMT datasets using LLaMA-68M and LLaMA-160M as draft models, and LLaMA-33B and Vicuna-33B as target models.}
\label{tab:alpha_baseline}
\end{table}

\begin{table*}[htb]
\footnotesize
\centering
\begin{tabular}{l|l|c|rrr|rrr}
    \toprule \multirow{2}*{Dataset}      & \multirow{2}*{Methods} & \multirow{2}*{Temp} & \multicolumn{3}{c|}{LLaMA-33B} & \multicolumn{3}{c}{Vicuna-33B} \\
                                         &                        &                     & $k$ Config.                    & Speedup                       & $\tau$        & $k$ Config. & Speedup               & $\tau$        \\
    \midrule \multirow{7}*{Alpaca}       & Baseline               & N/A                 & N/A                            & 1$\times$                     & 1             & N/A         & 1$\times$             & 1             \\
    \cmidrule{2-9}                       & SD (same $\gamma$)     & \multirow{4}*{0}    & 1x1x1x1x1                      & 2.71$\times$                  & 3.19          & 1x1x1x1     & 1.64$\times$          & 1.97          \\
                                         & SD (best $\gamma$)     &                     & {\tiny {1x1x1x1x1x1x1x1x1}}    & 2.89$\times$                  & 3.70          & 1x1x1x1x1x1 & 1.70$\times$          & 2.03          \\
                                         & Ours                   &                     & 4x2x2x1x1                      & \textbf{3.06$\times$}         & \textbf{3.93} & 8x2x1x1     & \textbf{2.06$\times$} & \textbf{2.56} \\
    \cmidrule{2-9}                       & SD (same $\gamma$)     & \multirow{3}*{1}    & 1x1x1x1                        & 1.69$\times$                  & 1.98          & 1x1x1       & 1.33$\times$          & 1.53          \\
                                         & SD (best $\gamma$)     &                     & 1x1x1x1x1                      & 1.73$\times$                  & 2.02          & 1x1x1x1x1   & 1.34$\times$          & 1.58          \\
                                         & Ours                   &                     & 8x2x1x1                        & \textbf{2.17$\times$}         & \textbf{2.71} & 16x1x1      & \textbf{1.73$\times$} & \textbf{2.10} \\
    \midrule \midrule \multirow{7}*{WMT} & Baseline               & N/A                 & N/A                            & 1$\times$                     & 1             & N/A         & 1$\times$             & 1             \\
    \cmidrule{2-9}                       & SD (same $\gamma$)     & \multirow{4}*{0}    & 1x1x1x1x1                      & 2.02$\times$                  & 2.38          & 1x1         & 1.15$\times$          & 1.29          \\
                                         & SD (best $\gamma$)     &                     & {\tiny 1x1x1x1x1x1x1}          & 2.06$\times$                  & 2.52          & 1x1         & 1.15$\times$          & 1.29          \\
                                         & Ours                   &                     & 4x2x2x1x1                      & \textbf{2.24$\times$}         & \textbf{2.87} & 16x2       & \textbf{1.45$\times$} & \textbf{1.73} \\
    \cmidrule{2-9}                       & SD (same $\gamma$)     & \multirow{3}*{1}    & 1x1x1                          & 1.41$\times$                  & 1.63          & 1x1         & 1.13$\times$          & 1.26          \\
                                         & SD (best $\gamma$)     &                     & 1x1x1x1x1                      & 1.43$\times$                  & 1.69          & 1x1x1       & 1.14$\times$          & 1.31          \\
                                         & Ours                   &                     & 8x2x1                          & \textbf{1.72$\times$}         & \textbf{2.08} & 16x2        & \textbf{1.42$\times$} & \textbf{1.70} \\
    \bottomrule
\end{tabular}
\caption{Performance of each method on Alpaca and WMT datasets using LLaMA-68M as draft model, and LLaMA-33B and Vicuna-33B as target models.}
\label{tab:main_l68m_33b}
\end{table*}


\begin{table*}[!ht]
\footnotesize
\centering
\begin{tabular}{l|l|c|rrr|rrr}
    \toprule \multirow{2}*{Dataset}      & \multirow{2}*{Methods} & \multirow{2}*{Temp} & \multicolumn{3}{c|}{LLaMA-33B} & \multicolumn{3}{c}{Vicuna-33B} \\
                                         &                        &                     & $k$ Config.                    & Speedup                       & $\tau$        & $k$ Config. & Speedup               & $\tau$        \\
    \midrule \multirow{7}*{Alpaca}       & Baseline               & N/A                 & N/A                            & 1$\times$                     & 1             & N/A         & 1$\times$             & 1             \\
    \cmidrule{2-9}                       & SD (same $\gamma$)     & \multirow{4}*{0}    & 1x1x1x1                        & 2.10$\times$                  & 3.20          & 1x1x1       & 1.33$\times$          & 1.97          \\
                                         & SD (best $\gamma$)     &                     & 1x1x1x1x1                      & 2.16$\times$                  & 3.49          & 1x1         & 1.40$\times$          & 1.81          \\
                                         & Ours                   &                     & 4x2x2x1                        & \textbf{2.42$\times$}         & \textbf{3.87} & 8x2x2       & \textbf{1.72$\times$} & \textbf{2.59} \\
    \cmidrule{2-9}                       & SD (same $\gamma$)     & \multirow{3}*{1}    & 1x1x1                          & 1.43$\times$                  & 2.05          & 1x1         & 1.24$\times$          & 1.60          \\
                                         & SD (best $\gamma$)     &                     & 1x1                            & 1.47$\times$                  & 1.90          & 1x1         & 1.24$\times$          & 1.60          \\
                                         & Ours                   &                     & 8x2x2                          & \textbf{1.85$\times$}         & \textbf{2.79} & 16x2        & \textbf{1.62$\times$} & \textbf{2.21} \\
    \midrule \midrule \multirow{7}*{WMT} & Baseline               & N/A                 & N/A                            & 1$\times$                     & 1             & N/A         & 1$\times$             & 1             \\
    \cmidrule{2-9}                       & SD (same $\gamma$)     & \multirow{4}*{0}    & 1x1x1x1                        & 1.54$\times$                  & 2.42          & 1           & 1.06$\times$          & 1.25          \\
                                         & SD (best $\gamma$)     &                     & 1x1x1                          & 1.57$\times$                  & 2.25          & 1           & 1.06$\times$          & 1.25          \\
                                         & Ours                   &                     & 8x2x1x1                        & \textbf{1.81$\times$}         & \textbf{2.96} & 32          & \textbf{1.33$\times$} & \textbf{1.66} \\
    \cmidrule{2-9}                       & SD (same $\gamma$)     & \multirow{3}*{1}    & 1x1                            & 1.22$\times$                  & 1.62          & 1           & 1.06$\times$          & 1.25          \\
                                         & SD (best $\gamma$)     &                     & 1x1                            & 1.22$\times$                  & 1.62          & 1           & 1.06$\times$          & 1.25          \\
                                         & Ours                   &                     & 16x2                           & \textbf{1.52$\times$}         & \textbf{2.14} & 32          & \textbf{1.30$\times$} & \textbf{1.63} \\
    \bottomrule
\end{tabular}
\caption{Performance of each method on Alpaca and WMT datasets using LLaMA-160M as draft model, and LLaMA-33B and Vicuna-33B as target models.}
\label{tab:main_l160m_33b}
\vspace{-1ex}

\end{table*}

We begin by looking at the impact of different factors, such as the dataset, supervised fine-tuning, and sampling method, on acceptance rates.
As shown in Table~\ref{tab:alpha_baseline}, on the unfine-tuned target model, i.e., LLaMA, the draft model generates candidates with good acceptance rates if the inference is performed on the Alpaca dataset with argmax sampling. However, fine-tuning the target model (Vicuna, a fine-tuned version of LLaMA), changing the dataset\footnote{The discrepancy between the datasets can be attributed to the variation in prompt length: WMT averages 29 words per sample, compared to Alpaca's 9.} or the sampling method can lead to a decrease in acceptance rates. In most cases, we did not observe a significant effect of changing the target model on size on acceptance rates, while increasing the size of the draft model had a limited positive effect on acceptance rates.

Fig.~\ref{fig:alphas} illustrates the acceptance rates at different $k$ using LLaMA-68M as a draft model, and the results using LLaMA-160M as a draft model can be found in Appendix~\ref{sec:alphas_l160m}. As $k$ increases, we observe a consistent improvement in acceptance rates across the different models and datasets. The $\alpha$ curves tend to converge when $k$ exceeds 32, at which point it becomes difficult and uneconomical to increase $k$ further. These results demonstrate the effectiveness of our method in improving the acceptance rate with only a small value of $k$.

\subsection{Main Results}
\label{sec:main_results}

\begin{figure*}[htb]
    \centering
    \includegraphics[scale=0.39]{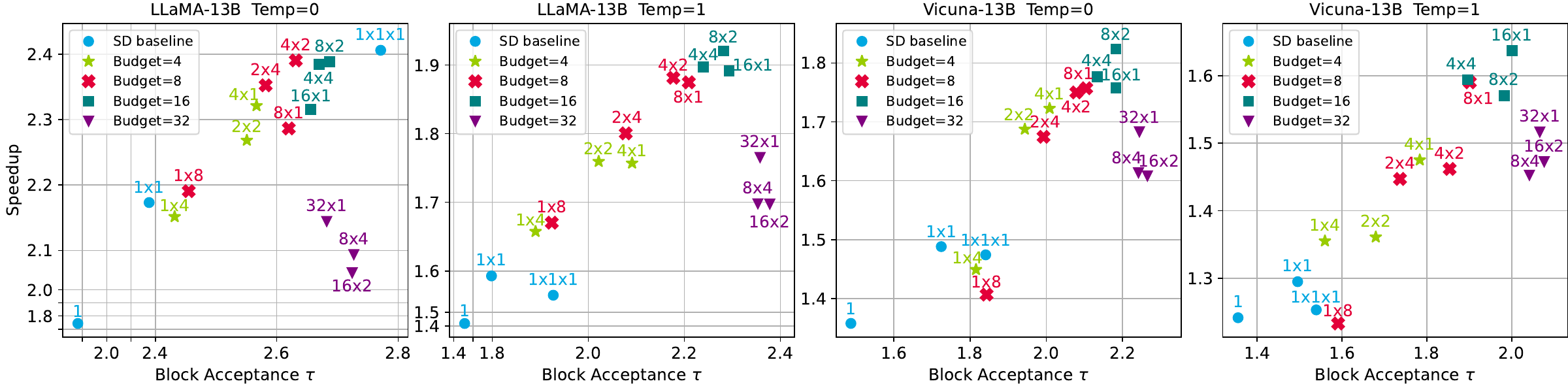}
    \vspace{-1ex}
    \caption{Speedup and block efficiency ($\tau$) for different $k$ configurations, where the dataset is Alpaca, using LLaMA-68M as the draft model.}
    \vspace{-1ex}
    \label{fig:budget_config}
\end{figure*}

\begin{table*}
[htb] \footnotesize
\centering
\begin{tabular}{lc|rr|rr|rr|rr}
    \toprule \multirow{2}*{Algorithm} & \multirow{2}*{Tree Attn.} & \multicolumn{2}{c|}{LLaMA-13B} & \multicolumn{2}{c|}{Vicuna-13B} & \multicolumn{2}{c|}{LLaMA-33B} & \multicolumn{2}{c}{Vicuna-33B} \\
                                      &                           & Speedup                        & $\tau$                         & Speedup                        & $\tau$                        & Speedup & $\tau$ & Speedup & $\tau$ \\
    \midrule
    MCSS w/o Replacement.    & \Checkmark   & 1.91$\times$     & 2.45        & 1.64$\times$     & 2.00 & 2.17$\times$     & 2.71        & 1.73$\times$     & 2.10   \\
    MCSS w/ Replacement.     & \Checkmark   & 1.89$\times$     & 2.39        & 1.55$\times$     & 1.95 & 2.04$\times$     & 2.55        & 1.66$\times$     & 2.01   \\
    MCNS                     & \Checkmark   & 1.21$\times$     & 1.52        & 1.45$\times$     & 1.78 & 1.33$\times$     & 1.65        & 1.50$\times$     & 1.80   \\
    MCSS w/o Replacement.    & \XSolidBrush & 1.61$\times$     & 2.47        & 1.36$\times$     & 1.99 & 1.84$\times$     & 2.68        & 1.51$\times$     & 2.09   \\
    \bottomrule
\end{tabular}
\caption{Ablation experiments on Tree Attention and different verification algorithms, where the dataset is Alpaca, the draft model is LLaMA-68M, and temp = 1 (standard sampling).}
\label{tab:ablation}
\vspace{-1ex}
\end{table*}

Given a draft of size $\gamma$, our method samples multiple tokens at each step, assuming they are $k_1, k_1, \cdots, k_\gamma$ respectively, generating a total of $K=\prod_{i=1}^\gamma k_i$ candidates. This constitutes a huge search space of hyperparameter. For efficiency reasons, we restrict the total budget $K$ to a maximum of 32, and $k_i\in \{1,2,4,8,16,32\}$. For the performance of our method under different hyperparameters, see Section~\ref{sec:budget_config}. For each setting, we report the performance of our method for the best combination of $k$s, as well as the SD with the same $\gamma$ and the SD with the best $\gamma$. We place the results of 13B versions in Appendix~\ref{sec:main_results_13b} to save space.

Table~\ref{tab:main_l68m_33b} shows the performance of each method using LLaMA-68m as the draft model. SD accelerates the LLaMA-33b model well on the Alpca dataset with argmax sampling, while the boost of our method is limited. However, when we observe a degradation in acceptance rate, as seen when using Vicuna as the target model, employing standard sampling, or working with the WMT dataset, the performance of SD degrades significantly. In these scenarios, our method demonstrates considerable improvement over SD. Furthermore, the difference observed between SD (same $\gamma$) and SD (best $\gamma$) validates our claim that further increasing $\gamma$ does not yield significant gains.

Replacing the draft model with LLaMA-160m, the results are shown in Table~\ref{tab:main_l160m_33b}. The notable difference from LLaMA-68m is that the latency of LLaMA-160m is much higher. Consequently, the higher cost of the invocations leads to a general shrinkage of $\gamma$. Our method achieves a similar improvement as in Table~\ref{tab:main_l68m_33b}, and in some cases, our method can works at a larger $\gamma$ compared to SD~(best $\gamma$), because it compensates for the additional overhead of the draft model by improving the acceptance rate.

Overall, our approach consistently achieves higher speedup and block efficiency compared to and SD baseline, demonstrating the effectiveness in improving the efficiency of the target model.

\subsection{Budget Configuration}
\label{sec:budget_config}

In this section, we examine the performance variations under different budget configurations. Fig.~\ref{fig:budget_config} shows the performance of the 13B-sized target model with different $k$ configurations on the Alpaca dataset, where we fixed $\gamma=2$ for clarity.

Our analysis yields three key insights.
Firstly, the monotonically decreasing configuration always outperforms the monotonically increasing configuration for the same budget (monotonic rule), e.g., 4x2 outperforms 2x4. This is because the acceptance of the next token is governed by the acceptance of the preceding tokens. Therefore, it is a natural strategy to use a monotonically decreasing configuration to preferentially improve the acceptance of the preceding tokens. The monotonic rule is practically useful in reducing the overhead in hyperparameter search.

Secondly, configurations with the same budget have roughly the same performance if we follow the monotonic rule. For instance, the 16-budget group is roughly centrally distributed, as is the 32-budget group. This underscores the robustness of our approach given a specific budget.

Lastly, despite a higher block efficiency, the 32-budget group demonstrates a lower speedup than the 16-budget group. This counterintuitive outcome stems from the diminishing returns of block efficiency gains as budget increases, which fail to compensate for the latency inherent to larger budgets.

\subsection{Ablation Study}

We conducted an ablation study to investigate the impact of our proposed improvements on performance, given optimal $k$ configurations as in Section~\ref{sec:main_results}. Our focus was on Tree Attention and verification algorithms that handle multiple candidates. More specifically, we looked at multi-candidate speculative sampling (MCSS) both with and without replacement, as well as a multi-candidate variant of naive sampling (MCNS)\footnote{MCNS first sample a token $x$ from $p(x)$ and accepts $\tilde{x}$ if $x\in \{\tilde{x}^1, \cdots, \tilde{x}^k\}$, otherwise it rejects the candidates and returns $x$.}. The experiments are based on standard sampling. For argmax sampling, MCSS with replacement degenerates to SD, as sampling with $\text{temp}=0$ consistently produces the top1 token. 
Additionally, it can be conclusively demonstrated that MCSS without replacement is equivalent to MCNS when using argmax sampling.

The findings, as presented in Table~\ref{tab:ablation}, reveal that MCSS offers the most significant improvement when compared to MCNS. Tree Attention also contributes a crucial role, not altering the expected block efficiency but substantially reducing the communication overhead.
The improvement brought by MCSS without replacement is limited, probably because repeated sampling is not as likely to happen when $k_i$ is small.

\section{Method Generalization Across Models}

\begin{table}[htb]
\footnotesize
\centering
\begin{tabular}{llcc}
    \toprule
    Draft model            & Target model   & Temp                  & $\Delta \alpha$                    \\
    \midrule \multirow{4}*{\scriptsize LLaMA-68M}  & {\scriptsize LLaMA-13B}        & 0 & 0.76 $\rightarrow$ 0.88 \\
                                                   & {\scriptsize Vicuna-13B}       & 0 & 0.49 $\rightarrow$ 0.67 \\
                                                   & {\scriptsize LLaMA2-13B}       & 0 & 0.75 $\rightarrow$ 0.88 \\
                                                   & {\scriptsize LLaMA2-13B-chat}  & 0 & 0.47 $\rightarrow$ 0.66 \\
    \midrule
    \multirow{4}*{\scriptsize Vicuna-68M}          & {\scriptsize LLaMA-13B}        & 0 & 0.75 $\rightarrow$ 0.90 \\
                                                   & {\scriptsize Vicuna-13B}       & 0 & 0.56 $\rightarrow$ 0.76 \\
                                                   & {\scriptsize LLaMA2-13B}       & 0 & 0.74 $\rightarrow$ 0.89 \\
                                                   & {\scriptsize LLaMA2-13B-chat}  & 0 & 0.55 $\rightarrow$ 0.75 \\
    \midrule
    \multirow{3}*{\scriptsize OPT-125M}            & {\scriptsize OPT-13B}          & 0 & 0.86 $\rightarrow$ 0.95 \\
                                                   & {\scriptsize OPT-30B}          & 0 & 0.83 $\rightarrow$ 0.94 \\
                                                   & {\scriptsize OPT-iml-30B}      & 0 & 0.81 $\rightarrow$ 0.93 \\
    \midrule
    \midrule \multirow{4}*{\scriptsize LLaMA-68M}  & {\scriptsize LLaMA-13B}        & 1 & 0.51 $\rightarrow$ 0.72 \\
                                                   & {\scriptsize Vicuna-13B}       & 1 & 0.35 $\rightarrow$ 0.57 \\
                                                   & {\scriptsize LLaMA2-13B}       & 1 & 0.51 $\rightarrow$ 0.71 \\
                                                   & {\scriptsize LLaMA2-13B-chat}  & 1 & 0.35 $\rightarrow$ 0.57 \\
    \midrule
    \multirow{4}*{\scriptsize Vicuna-68M}          & {\scriptsize LLaMA-13B}        & 1 & 0.48 $\rightarrow$ 0.69 \\
                                                   & {\scriptsize Vicuna-13B}       & 1 & 0.46 $\rightarrow$ 0.68 \\
                                                   & {\scriptsize LLaMA2-13B}       & 1 & 0.49 $\rightarrow$ 0.69 \\
                                                   & {\scriptsize LLaMA2-13B-chat}  & 1 & 0.46 $\rightarrow$ 0.69 \\
    \midrule
    \multirow{3}*{\scriptsize OPT-125M}            & {\scriptsize OPT-13B}          & 1 & 0.63 $\rightarrow$ 0.81 \\
                                                   & {\scriptsize OPT-30B}          & 1 & 0.60 $\rightarrow$ 0.79 \\
                                                   & {\scriptsize OPT-iml-30B}      & 1 & 0.61 $\rightarrow$ 0.78 \\
    \bottomrule
\end{tabular}
\caption{Acceptance rate ($\alpha$) improvement from $k=1$ to $k=4$ on Alpaca dataset using various draft and target models.}
\label{tab:alpha_generalization}
\vspace{-1ex}

\end{table}

We evaluate the effectiveness of our method across a range of draft and target models.

For draft models, we explore the intuitive hypothesis that fine-tuning the draft model would enhance the alignment with target models. We also examine the compatibility of our method when applied to the fine-tuned draft models. To accomplish this, we use ShareGPT data to fine-tune the LLaMA-68M and LLaMA-160M models, following the training setup provided by the Vicuna suite~\cite{vicuna2023}. The resulting fine-tuned models are subsequently named Vicuna-68M and Vicuna-160M. With regard to the target models, our consideration extended to the LLaMA2 suite~\cite{touvron2023llama_b} and the OPT suite~\cite{zhang2022opt, iyer2022opt}.

We report the baseline acceptance rate at $k = 1$ and the acceptance rate at $k = 4$ for each model pair. Table~\ref{tab:alpha_generalization} shows the results for the Alpaca dataset, we leave the results for the other models and datasets in Appendix~\ref{sec:alpha_generalization_full}.
Based on the baseline acceptance rate, we do observe that the fine-tuning allows the draft model to align with Vicuna with little loss of alignment to LLaMA. The delta value suggests that our method is fully superimposable with fine-tuning, and brings even more improvement over the unfine-tuned draft models. Similarly, on both the LLaMA2 suite and the OPT suite, our method brings consistent improvements.


\section{Related Work}

The quest for enhancing the inference efficiency of deep neural networks encompasses a broad spectrum of strategies, including but not limited to distillation~\cite{hinton2015distilling}, quantisation~\cite{dettmers2022llm}, and sparcification~\cite{jaszczur2021sparse}. These techniques often introducing some level of loss. In contrast, speculative decoding, as introduced by \citealp{leviathan2023fast} and \citealp{chen2023accelerating}, effectively reduces the inference latency of LLMs without compromising model performance. Before them, blockwise parallel decoding~\cite{stern2018blockwise} speeds up the inference of autoregressive models based on a similar principle, but it only supports argmax decoding. 

Given the pivotal role of distributional consistency between the draft and target models, researchers have focused on aligning the draft model with the target model through additional training~\cite{miao2023specinfer, zhou2023distillspec, liu2023online}. However, our empirical findings suggest that this alignment is less robust on out-of-distribution data (WMT) compared to in-distribution data (Alpaca).
In line with our research, some studies have employed multiple candidates to improve the acceptance rate.
\citealp{miao2023specinfer} utilises multiple draft models to generate diverse candidates, while \citealp{medusa} trains additional prediction heads for the same purpose. Their work also incorporates tree attention to reduce the communication overhead associated with multiple candidates.
Similar to our approach, \citealp{sun2023spectr} also sample multiple candidates from the draft model. The difference is that they derive the algorithm for multi-candidate verification from the perspective of optimal transport, which requires linear programming for its implementation.


\section{Conclusion}

This paper introduces multi-candidate speculative decoding. This method leverages the full potential of multiple candidates generated by the draft model, thereby improving the acceptance rate without compromising the output quality of the target model. We further augment this approach with Tree Attention that reduces communication overhead. Extensive testing across various models, decoding settings, and datasets has shown that our method consistently reduces latency compared to standard speculative decoding. Our method works out-of-the-box and also benefits from works that improve acceptance rates with additional training.



\bibliography{anthology,custom}

\appendix


\section{Proof of Multi-Candidate Speculative Sampling}
\label{sec:proof_MCSS}

Given any distributions $p(x)$ and $q(x)$, we now prove that the token returned from Algorithm~\ref{algorithm:MCSS} are distributed identically to those sampled from $p(x)$ alone.

\begin{proof}
Let $\tilde{x}_1,\cdots,\tilde{x}_k$ be the candidate tokens sampled from $q(x)$. We define
\begin{equation}
\label{equ:def_event_r}
\begin{split}
&\mathcal{R}_0:=\emptyset,\\
&\mathcal{R}_n:=\text{Reject }\tilde{x}_1,\cdots,\tilde{x}_n,
\end{split}
\end{equation}
and
\begin{equation}
\begin{split}
&\hat{p}_0(x):=p(x),\\
&\hat{p}_n(x):=\frac{\max(\hat{p}_{n-1}(x)-q(x),0)}{\sum_x {\max(\hat{p}_{n-1}(x)-q(x), 0)}}.
\end{split}
\end{equation}
According to Algorithm~\ref{algorithm:MCSS}, there is
\begin{equation}
\small
P(\text{Accept }\tilde{x}_n|\tilde{x}_n=x, \mathcal{R}_{n-1})=\min(1, \frac{\hat{p}_{n-1}(x)}{q(x)}).
\end{equation}
Thus we have
\begin{equation}
\small
\begin{split}
&P(x, \text{Accept }\tilde{x}_n|\mathcal{R}_{n-1})\\
&=P(\text{Accept }\tilde{x}_n,\tilde{x}_n=x|\mathcal{R}_{n-1})\\
&=P(\tilde{x}_n=x|\mathcal{R}_{n-1})P(\text{Accept }\tilde{x}_n|\tilde{x}_n=x, \mathcal{R}_{n-1})\\
&=P(\tilde{x}_n=x)P(\text{Accept }\tilde{x}_n|\tilde{x}_n=x, \mathcal{R}_{n-1})\\
&=q(x)\min(1, \frac{\hat{p}_{n-1}(x)}{q(x)})\\
&=\min(q(x),\hat{p}_{n-1}(x)).
\end{split}
\end{equation}
The probability of rejecting $\tilde{x}_n$ is
\begin{equation}
\label{equ:reject_xn}
\begin{split}
&P(\text{Reject }\tilde{x}_n|\mathcal{R}_{n-1})=1-P(\text{Accept }\tilde{x}_n|\mathcal{R}_{n-1})\\
&=1-\sum_xP(\text{Accept }\tilde{x}_n,\tilde{x}_n=x|\mathcal{R}_{n-1})\\
&=1-\sum_x\min(q(x),\hat{p}_{n-1}(x))\\
&=\sum_x\max(\hat{p}_{n-1}(x)-q(x),0).
\end{split}
\end{equation}
Let $A_n:=P(x| \mathcal{R}_{n})$. We have
\begin{equation}
\small
\begin{split}
&A_n = P(x| \mathcal{R}_{n})\\
&= P(x, \text{Accept }\tilde{x}_{n+1} | \mathcal{R}_{n}) + P(x, \text{Reject }\tilde{x}_{n+1} | \mathcal{R}_{n})\\
&= P(x, \text{Accept }\tilde{x}_{n+1} | \mathcal{R}_{n}) \\
&\qquad + P(\text{Reject }\tilde{x}_{n+1}| \mathcal{R}_{n}) P(x| \mathcal{R}_{n+1})\\
&=q(x)\min(1, \frac{\hat{p}_n(x)}{q(x)})+\sum_x\max(\hat{p}_n(x)-q(x),0)A_{n+1}\\
&=\min(q(x), \hat{p}_n(x))+\sum_x\max(\hat{p}_n(x)-q(x),0)A_{n+1}.
\end{split}
\end{equation}
According to Algorithm~\ref{algorithm:MCSS}, $A_k = P(x| \mathcal{R}_{k})=\hat{p}_k(x)$. So we get the recursive formula
\begin{equation}
\small
\begin{split}
&A_{k-1}=\min(q(x), \hat{p}_{k-1}(x))\\
&\qquad +\sum_x\max(\hat{p}_{k-1}(x)-q(x),0)A_k\\
&=\min(q(x), \hat{p}_{k-1}(x))\\
&\qquad +\sum_x\max(\hat{p}_{k-1}(x)-q(x),0)\hat{p}_k(x)\\
&=\min(q(x), \hat{p}_{k-1}(x))\\
&+\sum_x\max(\hat{p}_{k-1}(x)-q(x),0)\frac{\max(\hat{p}_{k-1}(x)-q(x),0)}{\sum_x {\max(\hat{p}_{k-1}(x)-q(x), 0)}}\\
&=\min(q(x), \hat{p}_{k-1}(x))+\max(\hat{p}_{k-1}(x)-q(x),0)\\
&=\hat{p}_{k-1}(x).
\end{split}
\end{equation}
Iteratively, we have
\begin{equation}
\begin{split}
A_0&=\hat{p}_{0}(x)=p(x),\\
P(x)&=P(x| \mathcal{R}_{0})=A_0=p(x).
\end{split}
\end{equation}
\end{proof}

\section{Proof of Multi-Candidate Speculative Sampling without Replacement}
\label{sec:proof_MCSSwoR}

Given any distributions $p(x)$ and $q(x)$, we now prove that the token returned from Algorithm~\ref{algorithm:MCSSwoR} are distributed identically to those sampled from $p(x)$ alone.

\begin{proof}
Let $\tilde{x}_1,\cdots,\tilde{x}_k$ be the candidate tokens sampled without replace from $q$, as in Eq.~(\ref{equ:q_bar}).
We define
\begin{equation}
\begin{split}
&\bar{p}_0(x)=p(x),\\
&\bar{p}_n(x)=\frac{\max(\bar{p}_{n-1}(x)-\bar{q}_{n-1}(x),0)}{\sum_x\max(\bar{p}_{n-1}(x)-\bar{q}_{n-1}(x),0)}.
\end{split}
\end{equation}
According to Algorithm~\ref{algorithm:MCSSwoR}, there is
\begin{equation}
\begin{split}
P(\text{Accept }\tilde{x}_{n+1}|\mathcal{R}_n,\tilde{x}_1,\cdots,\tilde{x}_{n+1})\\
=\min(1,\frac{\bar{p}_n(\tilde{x}_{n+1})}{\bar{q}_n(\tilde{x}_{n+1})}).
\end{split}
\end{equation}
Here we reuse the definition of $\mathcal{R}$ from Eq.~(\ref{equ:def_event_r}). 
Thus we have
\begin{equation}
\small
\begin{split}
&P(x, \text{Accept }\tilde{x}_{n+1}|\mathcal{R}_n,\tilde{x}_1,\cdots,\tilde{x}_n)\\
&=P(\tilde{x}_{n+1}=x, \text{Accept }\tilde{x}_{n+1}|\mathcal{R}_n,\tilde{x}_1,\cdots,\tilde{x}_n)\\
&=P(\tilde{x}_{n+1}=x|\mathcal{R}_n,\tilde{x}_1,\cdots,\tilde{x}_n)P(\text{Accept}\tilde{x}_{n+1}|\mathcal{R}_n,\tilde{x}_1,\cdots,\tilde{x}_{n+1})\\
&=\bar{q}_n(x)\min(1,\frac{\bar{p}_n(x)}{\bar{q}_n(x)})\\
&=\min(\bar{q}_n(x),\bar{p}_n(x)).
\end{split}
\end{equation}
and
\begin{equation}
\small
\begin{split}
&P(x, \text{Reject }\tilde{x}_{n+1}|\mathcal{R}_n,\tilde{x}_1,\cdots,\tilde{x}_n)\\
&=\sum_{\tilde{x}_{n+1}}P(x, \text{Reject }\tilde{x}_{n+1},\tilde{x}_{n+1}|\mathcal{R}_n,\tilde{x}_1,\cdots,\tilde{x}_n)\\
&=\sum_{\tilde{x}_{n+1}}[P(\tilde{x}_{n+1}|\mathcal{R}_n,\tilde{x}_1,\cdots,\tilde{x}_n)\\
&\qquad \times P(x,\text{Reject }\tilde{x}_{n+1}|\mathcal{R}_n,\tilde{x}_1,\cdots,\tilde{x}_{n+1})]\\
&=\sum_{\tilde{x}_{n+1}}[P(\tilde{x}_{n+1}|\mathcal{R}_n,\tilde{x}_1,\cdots,\tilde{x}_n)\\
&\qquad \times P(\text{Reject }\tilde{x}_{n+1}|\mathcal{R}_n,\tilde{x}_1,\cdots,\tilde{x}_{n+1})\\
&\qquad \times P(x|\mathcal{R}_{n+1},\tilde{x}_1,\cdots,\tilde{x}_{n+1})]\\
&=\sum_{\tilde{x}_{n+1}}[P(\tilde{x}_{n+1}|\mathcal{R}_n,\tilde{x}_1,\cdots,\tilde{x}_n)\\
&\qquad \times (1-P(\text{Accept }\tilde{x}_{n+1}|\mathcal{R}_n,\tilde{x}_1,\cdots,\tilde{x}_{n+1}))\\
&\qquad \times P(x|\mathcal{R}_{n+1},\tilde{x}_1,\cdots,\tilde{x}_{n+1})]\\
&=\sum_{\tilde{x}_{n+1}}[\bar{q}_n(\tilde{x}_{n+1})(1-\min(1,\frac{\bar{p}_n(\tilde{x}_{n+1})}{\bar{q}_n(\tilde{x}_{n+1})}))\\
&\qquad \times P(x|\mathcal{R}_{n+1},\tilde{x}_1,\cdots,\tilde{x}_{n+1})]\\
&=\sum_{\tilde{x}_{n+1}}\{[\bar{q}_n(\tilde{x}_{n+1})-\min(\bar{q}_n(\tilde{x}_{n+1}),\bar{p}_n(\tilde{x}_{n+1}))]\\
&\qquad \times P(x|\mathcal{R}_{n+1},\tilde{x}_1,\cdots,\tilde{x}_{n+1})\}.
\end{split}
\end{equation}
Let $A_n:=P(x|\mathcal{R}_n,\tilde{x}_1,\cdots,\tilde{x}_n)$. We have
\begin{equation}
\small
\begin{split}
&A_n=P(x|\mathcal{R}_n,\tilde{x}_1,\cdots,\tilde{x}_n)\\
&=P(x, \text{Accept }\tilde{x}_{n+1}|\mathcal{R}_n,\tilde{x}_1,\cdots,\tilde{x}_n)\\
&\qquad +P(x, \text{Reject }\tilde{x}_{n+1}|\mathcal{R}_n,\tilde{x}_1,\cdots,\tilde{x}_n)\\
&=\min(\bar{q}_n(x),\bar{p}_n(x))\\
&+\sum_{\tilde{x}_{n+1}}\{[\bar{q}_n(\tilde{x}_{n+1})-\min(\bar{q}_n(\tilde{x}_{n+1}),\bar{p}_n(\tilde{x}_{n+1}))]A_{n+1}\}.
\end{split}
\end{equation}
Because $\bar{q}_n(x)$ is independent of $\tilde{x}_{n+1}$. Thus $\bar{p}_n$ is independent of $x_n$ holds for $n = 1,\cdots,k$. 
According to Algorithm~\ref{algorithm:MCSSwoR},
\begin{equation}
\begin{split}
A_k &= P(x|\mathcal{R}_k,\tilde{x}_1,\cdots,\tilde{x}_k)=\bar{p}_k(x),
\end{split}
\end{equation}
so $A_k$ is also independent of $\tilde{x}_{k}$. We have
\begin{equation}
\small
\begin{split}
&A_{k-1}\\
&=\min(\bar{q}_{k-1}(x),\bar{p}_{k-1}(x))\\
& +\sum_{\tilde{x}_{k}}\{[\bar{q}_{k-1}(\tilde{x}_{k})-\min(\bar{q}_{k-1}(\tilde{x}_{k}),\bar{p}_{k-1}(\tilde{x}_{k}))]A_{k}\}\\
&=\min(\bar{q}_{k-1}(x),\bar{p}_{k-1}(x))\\
&+A_{k}\sum_{\tilde{x}_{k}}[\bar{q}_{k-1}(\tilde{x}_{k})-\min(\bar{q}_{k-1}(\tilde{x}_{k}),\bar{p}_{k-1}(\tilde{x}_{k}))]\\
&=\min(\bar{q}_{k-1}(x),\bar{p}_{k-1}(x))\\
&+A_{k}[\sum_{\tilde{x}_{k}}\bar{q}_{k-1}(\tilde{x}_{k})-\sum_{\tilde{x}_{k}}\min(\bar{q}_{k-1}(\tilde{x}_{k}),\bar{p}_{k-1}(\tilde{x}_{k}))]\\
&=\min(\bar{q}_{k-1}(x),\bar{p}_{k-1}(x))\\
&+A_{k}[\sum_{\tilde{x}_{k}}\bar{p}_{k-1}(\tilde{x}_{k})-\sum_{\tilde{x}_{k}}\min(\bar{q}_{k-1}(\tilde{x}_{k}),\bar{p}_{k-1}(\tilde{x}_{k}))]\\
&=\min(\bar{q}_{k-1}(x),\bar{p}_{k-1}(x))\\
&+A_{k}\sum_{\tilde{x}_{k}}[\bar{p}_{k-1}(\tilde{x}_{k})-\min(\bar{q}_{k-1}(\tilde{x}_{k}),\bar{p}_{k-1}(\tilde{x}_{k}))]\\
&=\min(\bar{q}_{k-1}(x),\bar{p}_{k-1}(x))\\
&+A_{k}\sum_{\tilde{x}_{k}}\max(\bar{p}_{k-1}(\tilde{x}_{k})-\bar{q}_{k-1}(\tilde{x}_{k}),0)\\
&=\min(\bar{q}_{k-1}(x),\bar{p}_{k-1}(x))\\
&+[\frac{\max(\bar{p}_{k-1}(x)-\bar{q}_{k-1}(x),0)}{\sum_x\max(\bar{p}_{k-1}(x)-\bar{q}_{k-1}(x),0)}\\
&\qquad\times\sum_{\tilde{x}_{k}}\max(\bar{p}_{k-1}(\tilde{x}_{k})-\bar{q}_{k-1}(\tilde{x}_{k}),0)]\\
&=\min(\bar{q}_{k-1}(x),\bar{p}_{k-1}(x))+\max(\bar{p}_{k-1}(x)-\bar{q}_{k-1}(x),0)\\
&=\bar{p}_{k-1}(x).
\end{split}
\end{equation}
Iteratively, we have
\begin{equation}
\begin{split}
A_0&=\bar{p}_{0}(x)=p(x),\\
P(x)&=P(x| \mathcal{R}_{0})=A_0=p(x).
\end{split}
\end{equation}
\end{proof}

\section{ Upper Bound on Acceptance Rate for Naive Sampling}
\label{sec:proof_NSUB}
We show that for any $k\in \mathcal{N}^+$, the acceptance rate of multi-candidate speculative sampling (with replacement) has an upper bound for multi-candidate naive sampling. Our proof references \citealp{miao2023specinfer}.

\begin{proof}
We use $P_S$ and $P_N$ to denote the probabilities involved in speculative sampling and naive sampling algorithms, respectively.

Since the acceptance rate is equal to $1-\sum_x P(x, \mathcal{R}_k)$, where we reuse the definition of $\mathcal{R}$ from Eq.~(\ref{equ:def_event_r}).
We now prove that $P_S(x, \mathcal{R}_k) \leq P_N(x, \mathcal{R}_k)$ always holds.

For speculative sampling, we have
\begin{equation}
\begin{split}
&P_S(x, \mathcal{R}_k)\\
&= P(x|\mathcal{R}_k)P(\mathcal{R}_k)\\
&=\hat{p}_k(x)\prod_{i=1}^{k}P(\text{Reject }\tilde{x}_i|\mathcal{R}_{i-1}).
\end{split}
\end{equation}
For naive sampling, there is
\begin{equation}
\begin{split}
P_N(x, \mathcal{R}_k) = p(x)(1-q(x))^k.
\end{split}
\end{equation}
If there is $\hat{p}_i(x)-q(x) \leq 0$, then $\hat{p}_j(x)=0$ holds for $j\geq i$. Thus $P_S(x, \mathcal{R}_k) = 0 \leq P_N(x, \mathcal{R}_k)$. Otherwise $\hat{p}_i(x)-q(x) > 0$ holds for $i=0,\cdots,k-1$, and we get
\begin{equation}
\begin{split}
\hat{p}_i(x)=\frac{\hat{p}_{i-1}(x)-q(x)}{\sum_x {\max(\hat{p}_{i-1}(x)-q(x), 0)}}
\end{split}
\end{equation}
By mathematical induction, if $k=1$, there is
\begin{equation}
\small
\begin{split}
&P_S(x, \mathcal{R}_1)=\hat{p}_1(x)P(\text{Reject }\tilde{x}_1|\mathcal{R}_{0})\\
&=\frac{\hat{p}_{0}(x)-q(x)}{\sum_x {\max(\hat{p}_{0}(x)-q(x), 0)}} \sum_x\max(\hat{p}_{0}(x)-q(x),0)\\
&=p(x)-q(x)\\
&\leq p(x)-p(x)q(x)\\
&=p(x)(1-q(x))\\
&=P_N(x, \mathcal{R}_1),
\end{split}
\end{equation}
where the value of $P(\text{Reject }\tilde{x}_1|\mathcal{R}_{0})$ comes from Eq.~(\ref{equ:reject_xn}).

Assume that $P_S(x, \mathcal{R}_k) \leq P_N(x, \mathcal{R}_k)$ holds for $k<n$, then
\begin{equation}
\small
\begin{split}
&P_S(x, \mathcal{R}_n)=\hat{p}_n(x)\prod_{i=1}^{n}P(\text{Reject }\tilde{x}_i|\mathcal{R}_{i-1})\\
&=\hat{p}_n(x)P(\text{Reject }\tilde{x}_{n}|\mathcal{R}_{n-1})\prod_{i=1}^{n-1}P(\text{Reject }\tilde{x}_i|\mathcal{R}_{i-1})\\
&=\frac{\hat{p}_{n-1}(x)-q(x)}{\sum_x {\max(\hat{p}_{n-1}(x)-q(x), 0)}} \sum_x\max(\hat{p}_{n-1}(x)-q(x),0)\\
&\qquad \times \prod_{i=1}^{n-1}P(\text{Reject }\tilde{x}_i|\mathcal{R}_{i-1})\\
&=[\hat{p}_{n-1}(x)-q(x)]\prod_{i=1}^{n-1}P(\text{Reject }\tilde{x}_i|\mathcal{R}_{i-1})\\
&\leq [\hat{p}_{n-1}(x)-\hat{p}_{n-1}(x)q(x)]\prod_{i=1}^{n-1}P(\text{Reject }\tilde{x}_i|\mathcal{R}_{i-1})\\
&= [1-q(x)]\hat{p}_{n-1}(x)\prod_{i=1}^{n-1}P(\text{Reject }\tilde{x}_i|\mathcal{R}_{i-1})\\
&\leq [1-q(x)]p(x)(1-q(x))^{n-1}\\
&\leq p(x)(1-q(x))^{n}\\
&=P_N(x, \mathcal{R}_n),
\end{split}
\end{equation}
\end{proof}

\begin{figure*}[htb]
    \centering
    \includegraphics[scale=0.39]{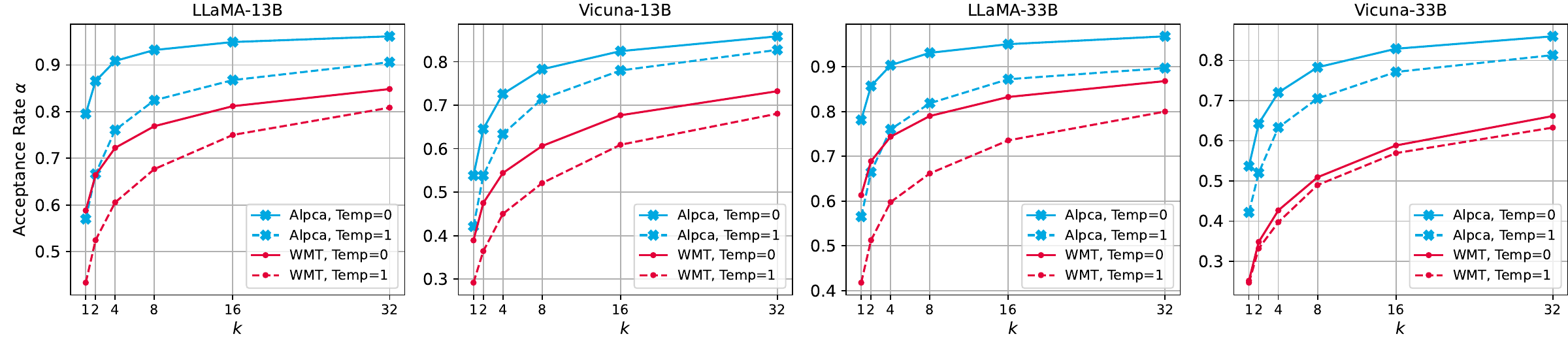}
    \vspace{-1ex}
    \caption{Acceptance rate ($\alpha$) curves given different $k$ using LLaMA-160M as draft model.}
    \vspace{-1ex}
    \label{fig:alphas_l160m}
\end{figure*}

\begin{table*}[htb]
\footnotesize
\centering
\begin{tabular}{l|l|c|rrr|rrr}
    \toprule \multirow{2}*{Dataset}      & \multirow{2}*{Methods} & \multirow{2}*{Temp} & \multicolumn{3}{c|}{LLaMA-13B} & \multicolumn{3}{c}{Vicuna-13B} \\
                                         &                        &                     & $k$ Config.                    & Speedup                       & $\tau$        & $k$ Config. & Speedup               & $\tau$        \\
    \midrule \multirow{7}*{Alpaca}       & Baseline               & N/A                 & N/A                            & 1$\times$                     & 1             & N/A         & 1$\times$             & 1             \\
    \cmidrule{2-9}                       & SD (same $\gamma$)     & \multirow{4}*{0}    & 1x1x1x1x1                      & 2.46$\times$                  & 3.35          & 1x1         & 1.49$\times$          & 1.72          \\
                                         & SD (best $\gamma$)     &                     & 1x1x1x1x1x1x1                  & 2.57$\times$                  & 3.72          & 1x1         & 1.49$\times$          & 1.72          \\
                                         & Ours                   &                     & 2x2x2x1x1                      & \textbf{2.75$\times$}         & \textbf{3.89} & 8x2         & \textbf{1.82$\times$} & \textbf{2.18} \\
    \cmidrule{2-9}                       & SD (same $\gamma$)     & \multirow{3}*{1}    & 1x1x1                          & 1.54$\times$                  & 1.93          & 1x1         & 1.30$\times$          & 1.50          \\
                                         & SD (best $\gamma$)     &                     & 1x1                            & 1.56$\times$                  & 1.79          & 1x1         & 1.30$\times$          & 1.50          \\
                                         & Ours                   &                     & 4x2x2                          & \textbf{1.91$\times$}         & \textbf{2.45} & 16x1        & \textbf{1.64$\times$} & \textbf{2.00} \\
    \midrule \midrule \multirow{7}*{WMT} & Baseline               & N/A                 & N/A                            & 1$\times$                     & 1             & N/A         & 1$\times$             & 1             \\
    \cmidrule{2-9}                       & SD (same $\gamma$)     & \multirow{4}*{0}    & 1x1x1                          & 1.69$\times$                  & 2.05          & 1x1         & 1.29$\times$          & 1.52          \\
                                         & SD (best $\gamma$)     &                     & 1x1x1                          & 1.69$\times$                  & 2.05          & 1x1x1x1     & 1.31$\times$          & 1.63          \\
                                         & Ours                   &                     & 4x2x1                          & \textbf{1.88$\times$}         & \textbf{2.37} & 8x1         & \textbf{1.54$\times$} & \textbf{1.85} \\
    \cmidrule{2-9}                       & SD (same $\gamma$)     & \multirow{3}*{1}    & 1x1                            & 1.30$\times$                  & 1.55          & 1x1         & 1.14$\times$          & 1.34          \\
                                         & SD (best $\gamma$)     &                     & 1x1x1                          & 1.31$\times$                  & 1.63          & 1           & 1.15$\times$          & 1.25          \\
                                         & Ours                   &                     & 8x2                            & \textbf{1.57$\times$}         & \textbf{1.95} & 16x1        & \textbf{1.39$\times$} & \textbf{1.72} \\
    \bottomrule
\end{tabular}
\caption{Performance of each method on Alpaca and WMT datasets using LLaMA-68M as draft model, and LLaMA-13B and Vicuna-13B as target models.}
\label{tab:main_l68m_13b}
\end{table*}

\begin{table*}[ht!]
\footnotesize
\centering
\begin{tabular}{l|l|c|rrr|rrr}
    \toprule \multirow{2}*{Dataset}      & \multirow{2}*{Methods} & \multirow{2}*{Temp} & \multicolumn{3}{c|}{LLaMA-13B} & \multicolumn{3}{c}{Vicuna-13B} \\
                                         &                        &                     & $k$ Config.                    & Speedup                       & $\tau$        & $k$ Config. & Speedup               & $\tau$        \\
    \midrule \multirow{7}*{Alpaca}       & Baseline               & N/A                 & N/A                            & 1$\times$                     & 1             & N/A         & 1$\times$             & 1             \\
    \cmidrule{2-9}                       & SD (same $\gamma$)     & \multirow{4}*{0}    & 1x1x1                          & 1.53$\times$                  & 2.91          & 1x1         & 1.15$\times$          & 1.83          \\
                                         & SD (best $\gamma$)     &                     & 1x1x1                          & 1.53$\times$                  & 2.91          & 1           & 1.16$\times$          & 1.54          \\
                                         & Ours                   &                     & 4x2x2                          & \textbf{1.70$\times$}         & \textbf{3.36} & 4x4         & \textbf{1.39$\times$} & \textbf{2.25} \\
    \cmidrule{2-9}                       & SD (same $\gamma$)     & \multirow{3}*{1}    & 1x1                            & 1.17$\times$                  & 1.90          & 1           & 1.09$\times$          & 1.42          \\
                                         & SD (best $\gamma$)     &                     & 1                              & 1.19$\times$                  & 1.57          & 1           & 1.09$\times$          & 1.42          \\
                                         & Ours                   &                     & 8x2                            & \textbf{1.41$\times$}         & \textbf{2.37} & 32          & \textbf{1.29$\times$} & \textbf{1.83} \\
    \midrule \midrule \multirow{7}*{WMT} & Baseline               & N/A                 & N/A                            & 1$\times$                     & 1             & N/A         & 1$\times$             & 1             \\
    \cmidrule{2-9}                       & SD (same $\gamma$)     & \multirow{4}*{0}    & 1x1                            & 1.18$\times$                  & 1.94          & 1           & 1.07$\times$          & 1.39          \\
                                         & SD (best $\gamma$)     &                     & 1x1                            & 1.18$\times$                  & 1.94          & 1           & 1.07$\times$          & 1.39          \\
                                         & Ours                   &                     & 16x1                           & \textbf{1.34$\times$}         & \textbf{2.31} & 16          & \textbf{1.25$\times$} & \textbf{1.68} \\
    \cmidrule{2-9}                       & SD (same $\gamma$)     & \multirow{3}*{1}    & 1                              & 1.04$\times$                  & 1.43          & 1           & 0.99$\times$          & 1.29          \\
                                         & SD (best $\gamma$)     &                     & 1                              & 1.04$\times$                  & 1.43          & 1           & 0.99$\times$          & 1.29          \\
                                         & Ours                   &                     & 32                             & \textbf{1.27$\times$}         & \textbf{1.81} & 32          & \textbf{1.20$\times$} & \textbf{1.68} \\
    \bottomrule
\end{tabular}
\caption{Performance of each method on Alpaca and WMT datasets using LLaMA-160M as draft model, and LLaMA-13B and Vicuna-13B as target models.}
\label{tab:main_l160m_13b}
\end{table*}

\section{Acceptance Rate Improvement}
\label{sec:alphas_l160m}

Fig.~\ref{fig:alphas_l160m} shows the acceptance rate improvement from increasing $k$ when using LLaMA-160M as a draft model.

\section{Main Results for 13B Models}
\label{sec:main_results_13b}

Table~\ref{tab:main_l68m_13b} and Table~\ref{tab:main_l160m_13b} shows the performance of each method when using target models of size 13B.

\section{Results Across Models}
\label{sec:alpha_generalization_full}

We present the improvements in acceptance rates for various model pairs under the Alpaca and WMT datasets in Table~\ref{tab:alpha_generalization_full}.
Our observe that the fine-tuned draft models, Vicuna-68M and Vicuna-160M, exhibit better alignment with both the Vicuna model and the LLaMA2-chat model, with a slight loss of alignment to the LLaMA base model and the LLaMA2 base model. Fine-tuning, however, shows domain-specific limitations, yielding less enhancement in baseline acceptance on the WMT dataset compared to the Alpaca dataset. Our method is versatile, proving compatible with both pre- and post-fine-tuning models. In the context of the OPT suite, our approach achieves peak enhancements in acceptance rates for the OPT-iml-30B model on the WMT dataset.

\begin{table*}[htb]
\footnotesize
\centering
\begin{tabular}{llcccc}
    \toprule
    \multirow{2}*{Draft model}            & \multirow{2}*{Target model}   & \multirow{2}*{Temp}                  & \multicolumn{2}{c}{$\Delta \alpha$}  \\
    & & &   Alpaca  &  WMT \\
    \midrule 
    \multirow{4}*{ LLaMA-68M}                      & { LLaMA-13B}        & 0 & 0.76 $\rightarrow$ 0.88 (+0.12)      & 0.55 $\rightarrow$ 0.68 (+0.13) \\
                                                   & { Vicuna-13B}       & 0 & 0.49 $\rightarrow$ 0.67 (+0.19)      & 0.36 $\rightarrow$ 0.51 (+0.14) \\
                                                   & { LLaMA2-13B}       & 0 & 0.75 $\rightarrow$ 0.88 (+0.13)      & 0.57 $\rightarrow$ 0.71 (+0.14) \\
                                                   & { LLaMA2-13B-chat}  & 0 & 0.47 $\rightarrow$ 0.66 (+0.19)      & 0.30 $\rightarrow$ 0.44 (+0.15) \\
    \midrule
    \multirow{4}*{ Vicuna-68M}                     & { LLaMA-13B}        & 0 & 0.75 $\rightarrow$ 0.90 (+0.14)      & 0.56 $\rightarrow$ 0.69 (+0.13) \\
                                                   & { Vicuna-13B}       & 0 & 0.56 $\rightarrow$ 0.76 (+0.20)      & 0.38 $\rightarrow$ 0.57 (+0.19) \\
                                                   & { LLaMA2-13B}       & 0 & 0.74 $\rightarrow$ 0.89 (+0.14)      & 0.56 $\rightarrow$ 0.69 (+0.13) \\
                                                   & { LLaMA2-13B-chat}  & 0 & 0.55 $\rightarrow$ 0.75 (+0.21)      & 0.32 $\rightarrow$ 0.53 (+0.21) \\
    \midrule
    \multirow{4}*{ LLaMA-160M}                     & { LLaMA-13B}        & 0 & 0.80 $\rightarrow$ 0.91 (+0.11)      & 0.59 $\rightarrow$ 0.72 (+0.13) \\
                                                   & { Vicuna-13B}       & 0 & 0.54 $\rightarrow$ 0.73 (+0.19)      & 0.39 $\rightarrow$ 0.54 (+0.15) \\
                                                   & { LLaMA2-13B}       & 0 & 0.78 $\rightarrow$ 0.90 (+0.12)      & 0.61 $\rightarrow$ 0.74 (+0.14) \\
                                                   & { LLaMA2-13B-chat}  & 0 & 0.52 $\rightarrow$ 0.72 (+0.19)      & 0.32 $\rightarrow$ 0.48 (+0.16) \\
    \midrule
    \multirow{4}*{ Vicuna-160M}                    & { LLaMA-13B}        & 0 & 0.78 $\rightarrow$ 0.91 (+0.12)      & 0.59 $\rightarrow$ 0.72 (+0.13) \\
                                                   & { Vicuna-13B}       & 0 & 0.62 $\rightarrow$ 0.81 (+0.19)      & 0.40 $\rightarrow$ 0.58 (+0.18) \\
                                                   & { LLaMA2-13B}       & 0 & 0.77 $\rightarrow$ 0.90 (+0.13)      & 0.59 $\rightarrow$ 0.72 (+0.13) \\
                                                   & { LLaMA2-13B-chat}  & 0 & 0.61 $\rightarrow$ 0.81 (+0.20)      & 0.33 $\rightarrow$ 0.54 (+0.22) \\
    \midrule
    \multirow{3}*{ OPT-125M}                       & { OPT-13B}          & 0 & 0.86 $\rightarrow$ 0.95 (+0.09)      & 0.97 $\rightarrow$ 0.99 (+0.02) \\
                                                   & { OPT-30B}          & 0 & 0.83 $\rightarrow$ 0.94 (+0.11)      & 0.80 $\rightarrow$ 0.89 (+0.09) \\
                                                   & { OPT-iml-30B}      & 0 & 0.81 $\rightarrow$ 0.93 (+0.12)      & 0.40 $\rightarrow$ 0.77 (+0.37) \\
    \midrule
    \midrule \multirow{4}*{ LLaMA-68M}             & { LLaMA-13B}        & 1 & 0.51 $\rightarrow$ 0.72 (+0.21)      & 0.39 $\rightarrow$ 0.56 (+0.17) \\
                                                   & { Vicuna-13B}       & 1 & 0.35 $\rightarrow$ 0.57 (+0.22)      & 0.25 $\rightarrow$ 0.41 (+0.16) \\
                                                   & { LLaMA2-13B}       & 1 & 0.51 $\rightarrow$ 0.71 (+0.20)      & 0.38 $\rightarrow$ 0.57 (+0.18) \\
                                                   & { LLaMA2-13B-chat}  & 1 & 0.35 $\rightarrow$ 0.57 (+0.22)      & 0.25 $\rightarrow$ 0.39 (+0.14) \\
    \midrule
    \multirow{4}*{ Vicuna-68M}                     & { LLaMA-13B}        & 1 & 0.48 $\rightarrow$ 0.69 (+0.21)      & 0.38 $\rightarrow$ 0.55 (+0.18) \\
                                                   & { Vicuna-13B}       & 1 & 0.46 $\rightarrow$ 0.68 (+0.22)      & 0.29 $\rightarrow$ 0.45 (+0.16) \\
                                                   & { LLaMA2-13B}       & 1 & 0.49 $\rightarrow$ 0.69 (+0.20)      & 0.38 $\rightarrow$ 0.56 (+0.18) \\
                                                   & { LLaMA2-13B-chat}  & 1 & 0.46 $\rightarrow$ 0.69 (+0.22)      & 0.30 $\rightarrow$ 0.49 (+0.18) \\
    \midrule \multirow{4}*{ LLaMA-160M}            & { LLaMA-13B}        & 1 & 0.57 $\rightarrow$ 0.76 (+0.19)      & 0.43 $\rightarrow$ 0.61 (+0.17) \\
                                                   & { Vicuna-13B}       & 1 & 0.42 $\rightarrow$ 0.63 (+0.21)      & 0.29 $\rightarrow$ 0.45 (+0.16) \\
                                                   & { LLaMA2-13B}       & 1 & 0.57 $\rightarrow$ 0.75 (+0.19)      & 0.43 $\rightarrow$ 0.61 (+0.18) \\
                                                   & { LLaMA2-13B-chat}  & 1 & 0.41 $\rightarrow$ 0.63 (+0.22)      & 0.29 $\rightarrow$ 0.44 (+0.15) \\
    \midrule
    \multirow{4}*{ Vicuna-160M}                    & { LLaMA-13B}        & 1 & 0.54 $\rightarrow$ 0.73 (+0.20)      & 0.41 $\rightarrow$ 0.58 (+0.18) \\
                                                   & { Vicuna-13B}       & 1 & 0.53 $\rightarrow$ 0.74 (+0.22)      & 0.31 $\rightarrow$ 0.48 (+0.18) \\
                                                   & { LLaMA2-13B}       & 1 & 0.54 $\rightarrow$ 0.73 (+0.19)      & 0.41 $\rightarrow$ 0.59 (+0.18) \\
                                                   & { LLaMA2-13B-chat}  & 1 & 0.54 $\rightarrow$ 0.75 (+0.22)      & 0.33 $\rightarrow$ 0.51 (+0.18) \\
    \midrule
    \multirow{3}*{ OPT-125M}                       & { OPT-13B}          & 1 & 0.63 $\rightarrow$ 0.81 (+0.18)      & 0.59 $\rightarrow$ 0.78 (+0.19) \\
                                                   & { OPT-30B}          & 1 & 0.60 $\rightarrow$ 0.79 (+0.18)      & 0.56 $\rightarrow$ 0.76 (+0.20) \\
                                                   & { OPT-iml-30B}      & 1 & 0.61 $\rightarrow$ 0.78 (+0.17)      & 0.44 $\rightarrow$ 0.68 (+0.24) \\
    \bottomrule
\end{tabular}
\caption{Acceptance rate ($\alpha$) improvement from $k=1$ to $k=4$ on Alpaca and WMT datasets using various draft and target models.}
\label{tab:alpha_generalization_full}
\vspace{-1ex}

\end{table*}

\end{document}